\documentclass[letterpaper,10pt,conference]{ieeeconf}
\IEEEoverridecommandlockouts
\usepackage{amsmath,amssymb}
\usepackage{booktabs}
\usepackage{graphicx}
\usepackage{microtype}
\usepackage{url}
\usepackage{balance}
\usepackage{float}
\usepackage{dblfloatfix}
\usepackage{placeins}
\title{Persistent Profile-Guided Reset-Free Recovery Improves Final Gait Selection after Irreversible Leg Damage}
\author{Minjoong Kim%
\thanks{Minjoong Kim is an independent researcher. \protect\\
Email: {\tt\small kimminjoong.mmm@gmail.com}.}}
\newcommand{\method}{PGRR}

\ifdefined
\fi
\begin{document}
\maketitle

\begin{abstract}
Damage-recovery algorithms often rank gaits from short trials, yet a gait that performs well briefly may not remain effective during continued execution. In reset-free recovery, each trial also changes the physical state inherited by the next. We introduce persistent profile-guided reset-free recovery (\method{}), which predicts multiple outcomes over repeated gait execution and uses conservative uncertainty-aware estimates to guide short online probes without resetting the robot. Across 108 matched MuJoCo tasks spanning leg amputation, distal fracture, and joint lock, \method{} improves final quality by 0.106 and slot-averaged recovery quality (SRQ) by 0.044 relative to an IT\&E-inspired Gaussian-process baseline, while reducing low-quality failures by 27.8 percentage points. Final-quality gains occur in all three damage classes. Against no adaptation, \method{} also improves final quality and reduces low-quality failures, whereas the aggregate SRQ difference is inconclusive. Development ablations identify conservative uncertainty handling as the clearest supported contributor to the observed gains. These results show that \method{} can identify stronger post-damage gaits under continuous physical state and that final-gait quality and recovery-summary performance provide complementary evidence for evaluating reset-free recovery.
\end{abstract}


\section{Introduction}
Robots enable monitoring and inspection in environments that are difficult or hazardous for people to access. Deployments in nuclear facilities and underwater survey and inspection illustrate the value of performing these tasks remotely or autonomously \cite{staniaszek2024,ribas2012}. As missions extend from individual inspections to repeated operation over days or weeks, maintaining mobility with limited human intervention becomes increasingly important \cite{staniaszek2024}. For a legged robot, structural damage creates a direct obstacle to continued operation because a gait developed for an intact body may no longer produce effective locomotion. Recovering useful motion with the remaining limbs is therefore an important step toward sustained autonomy \cite{bongard2006,cully2015}.

Post-damage adaptation has advanced through self-modeling, which revises a robot's body model after component loss \cite{bongard2006}, and Intelligent Trial and Error (IT\&E), which uses a library of behaviors computed offline to guide a small number of online tests \cite{cully2015}. Reset-free Trial and Error (RTE) further accounts for the physical state left by each preceding attempt \cite{chatzilygeroudis2018}. RTE evaluates recovery while successive trials inherit the state produced by earlier ones, but this does not by itself characterize how well the finally selected gait performs when executed separately after adaptation. This distinction matters because every attempted gait changes the robot's pose, velocity, and contacts. A search may eventually identify an effective gait even if some intermediate trials perform poorly, while a gait that appears promising during a short test may deteriorate during continued execution.

To address this distinction, we propose persistent profile-guided reset-free recovery (\method{}). In an offline stage, candidate gaits are repeatedly executed to learn how their performance changes beyond an initial short trial. These profiles describe survival, early-to-late locomotion quality, final quality, path-related performance, and energy use over continued execution. After damage, \method{} uses these profiles and their predictive uncertainty to select a promising gait for a short online probe. The observed performance is then used to update the predictions, while the best gait identified so far is executed between probes as the incumbent bridge. The robot is not reset between trials: each new probe starts from the physical state left by the preceding motion. In this way, \method{} uses short online tests to search for a gait that is expected to remain effective during continued execution while preserving the robot's physical state throughout recovery.

We implement \method{} on a simulated hexapod and evaluate it on a frozen holdout of 108 matched condition--seed tasks spanning leg amputation, distal fracture, and joint lock. Locomotion quality combines forward motion, path tracking, energy use, and transition stability. We report both SRQ, which summarizes recovery under a fixed scoring budget, and the quality of a separate 3-s execution of the selected gait. Relative to the tested IT\&E-inspired Gaussian-process baseline, \method{} improves final quality by 0.106 and SRQ by 0.044, while reducing low-quality failures by 27.8 percentage points. Final quality improves in all three damage classes.

This work makes three contributions. First, we distinguish final-gait quality from the overall recovery summary in reset-free damage recovery. Second, we introduce \method{}, which combines repeated-execution profiles, conservative uncertainty-aware selection, short online probes, and incumbent bridging without resetting the robot. Third, we evaluate the method on 108 matched MuJoCo tasks spanning leg amputation, distal fracture, and joint lock. \method{} improves final-gait quality and reduces low-quality failures, while SRQ gains depend on the comparison baseline. These results show that final performance and recovery-summary performance capture complementary aspects of reset-free recovery.

\section{Related Work}
Quality-diversity methods construct collections of distinct, high-performing behaviors \cite{mouret2015,pugh2016}. These collections allow a damaged robot to search among previously generated options rather than learn every movement from scratch. Work following IT\&E examines repertoire size in hexapods \cite{mailer2021}, chooses among libraries suited to different situations with APROL \cite{kaushik2020}, and uses hierarchical libraries for simulated and physical damage recovery \cite{allard2022,allard2023}. PGRR instead emphasizes how a candidate is expected to behave across repeated execution and how observations revise those expectations during recovery. Its experiments compare policies using a shared candidate pool and trial budget; the specific IT\&E-inspired implementation is defined in the experimental protocol.

Rapid motor adaptation adjusts locomotion to changing terrain and payload \cite{kumar2021}, whereas the present study concerns permanent loss or immobilization of leg components. Risk-sensitive methods also inform the choice of adaptation trials. Safe Bayesian optimization uses probabilistic performance models to seek improvement while enforcing safety constraints under stated assumptions \cite{sui2015,berkenkamp2023}. Safety-aware damage recovery can initialize these models with behavior predicted in simulation \cite{papaspyros2016}. PGRR uses uncertainty penalties and observed viability thresholds to select and extend trials, without claiming the formal safety guarantees of those methods. We examine both the quality summarized across recovery and the motion produced by the final selected controller in a separate execution interval.

\section{Problem and Evaluation}
\begin{figure*}[!tb]
\centering
\includegraphics[width=0.95\textwidth]
{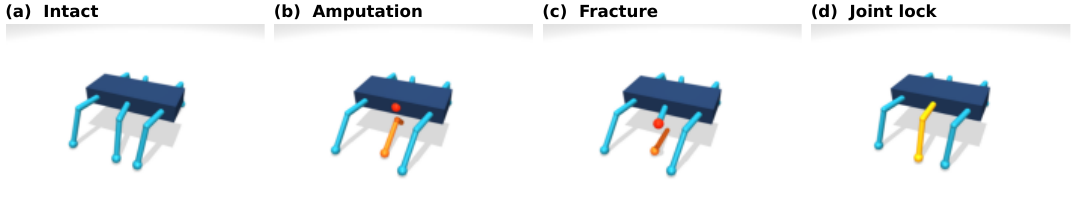}
\vspace{-15pt}
\caption{The four evaluated structural states at a common scale. Amputation removes a full leg, fracture removes its distal segment, and joint lock freezes both joints. Orange components are detached dynamic debris; yellow marks an immobilized leg. Illustrative renders use leg 1 at 0.30 s after damage and do not depict a measured recovery outcome.}
\label{fig:damage}
\end{figure*}
\subsection{Robot and damage}
Figure~\ref{fig:damage} shows the structural conditions. We simulate a hexapod with two actuated joints per leg in MuJoCo, using a 5-ms time step \cite{todorov2012}. Its gait lifts, advances, and lowers each foot before stance. Damage is applied at a prescribed time without resetting pose, velocity, or contacts. Amputation detaches a whole leg and disables its two actuators. Fracture detaches the distal segment and disables one actuator. Joint lock fixes both joints at their damage-time positions and disables their actuators. Detached segments remain as dynamic debris. These interventions model connectivity loss and joint immobilization, rather than material fracture or damage detection.

\subsection{Segment quality}
For each executed segment, we compare forward speed, lateral drift per forward distance, work per forward distance, and transition stability with an intact reference at matched friction. The corresponding scores $s,p,e,u$ are clipped to $[0,1]$. We define
\begin{equation}
q=(speu)^{1/4}.
\label{eq:quality}
\end{equation}
All scores are zero if the stable fraction is below 0.95. Otherwise, $s$ is the speed ratio; $p=(\ell_0+0.01)/(\ell+0.01)$, where $\ell$ is maximum lateral drift divided by forward distance; and $e$ is the reference-to-measured ratio of work per distance. Distance is floored at 0.005 m. The stability score is the transition stable fraction multiplied by the clipped ratio $(\theta_0+2)/(\theta+2)$, where $\theta$ is maximum transition tilt in degrees. Subscript 0 denotes the intact reference. The geometric mean penalizes a poor component even when forward speed is high.

\subsection{Recovery summary and final performance}
An adaptive trial contains a 0.75-s probe, an optional 0.75-s extension, and a 0.75-s bridge using the current incumbent. Let $\bar q_k$ be the mean quality of the segments actually executed in trial $k$, $K\leq B$ the number of trials, and $q_F$ the quality of a separate 3-s final execution. With $B=10$, slot-averaged recovery quality (SRQ) is
\begin{equation}
\mathrm{SRQ}=\frac{\sum_{k=1}^{K}\bar q_k+(B-K+1)q_F}{B+1}.
\label{eq:srq}
\end{equation}
Unused trial slots are filled with $q_F$ after early stopping. This padding is a scoring convention, not an additional physical measurement. SRQ weights slots equally despite differing durations; it is not a time integral. Because it includes $q_F$, SRQ and final quality are related outcomes rather than independent evidence. We also report the low-quality failure rate, defined by $q_F\leq0.05$. This threshold detects very poor locomotion and is not a separate physical-collapse detector.

\section{Profile-Guided Recovery}
\begin{figure*}[!tb]
\centering
\includegraphics[width=0.95\textwidth]
{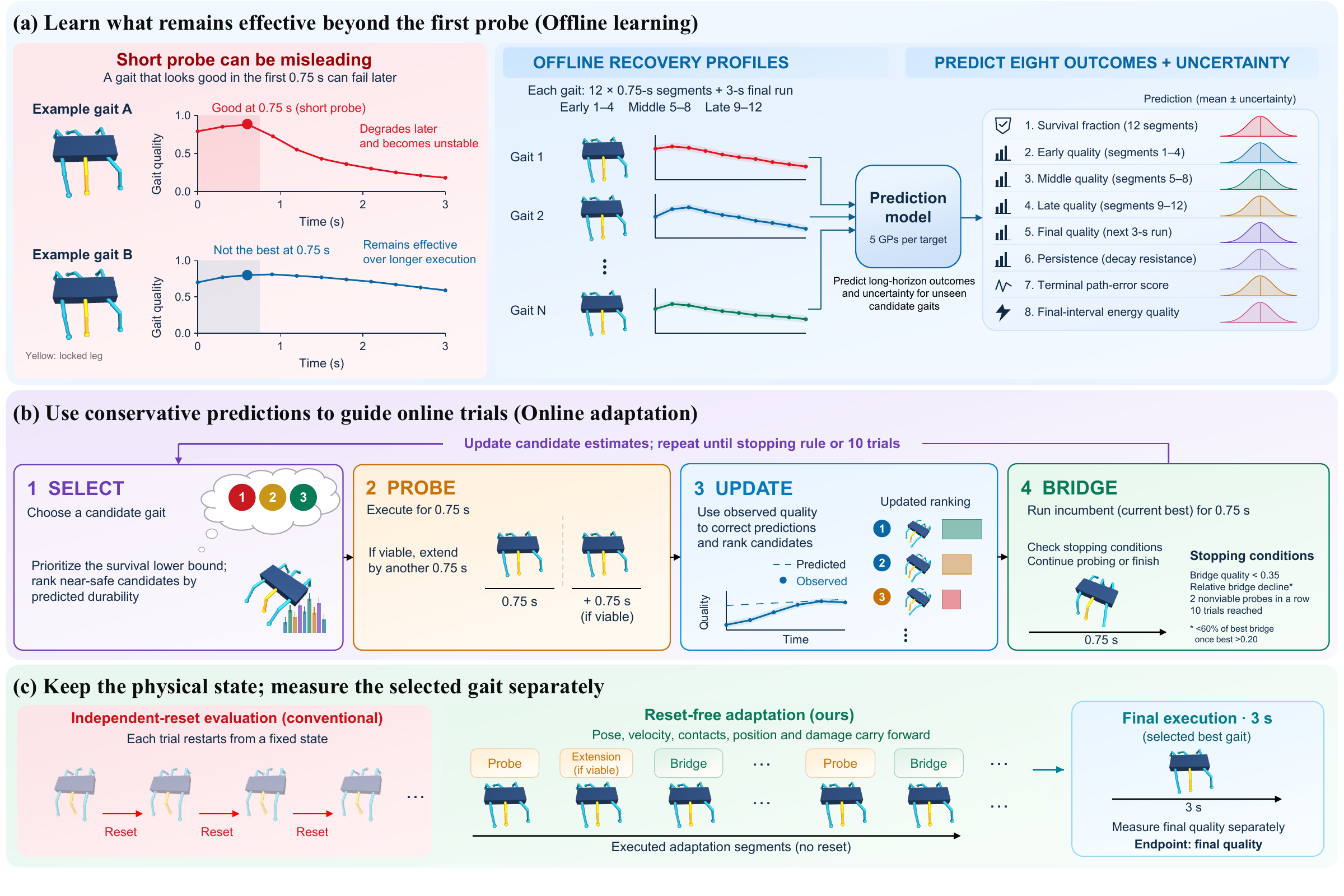}
\vspace{-15pt}
\caption{PGRR from offline profiling to continuous online execution. (a) Repeated execution reveals behavior that a short probe can miss; twelve 0.75-s segments followed by a 3-s endpoint interval supply eight prediction targets. (b) Conservative selection, viability-gated probes, online correction, and incumbent bridges repeat until a stopping condition or the ten-trial budget. (c) Conventional independent-reset evaluation is shown only as a conceptual contrast. PGRR carries pose, velocity, contacts, position, and damage through adaptation and into the separately scored 3-s final interval. Robot snapshots are rendered from MuJoCo development runs; yellow marks the locked leg. Curves and timelines are schematic.}
\label{fig:method}
\end{figure*}

Figure~\ref{fig:method} presents the method in execution order. Panel A describes the offline stage. Each candidate gait is executed repeatedly to form a long-horizon profile, and five GP models per target predict eight outcomes and their uncertainty for candidate gaits. Panel B describes online adaptation after damage. The policy selects a candidate conservatively, executes a 0.75-s probe, extends viable probes by another 0.75 s, uses the observation to correct its predictions, and runs the current incumbent for 0.75 s as a bridge. This select--probe--update--bridge cycle continues until a stopping condition or the ten-trial budget is reached.

Panel C distinguishes the physical trajectory from the reported endpoint. No trial restarts from a common initial state: each probe inherits the pose, velocity, contacts, position, and damage state left by the preceding motion. The bridge is therefore an executed movement that changes the next probe's initial condition rather than a reset. After adaptation, the selected gait runs for 3 s from the current state. This interval is measured separately as final quality even though the physical state remains continuous.
\subsection{Candidate gaits and offline profiles}
Each gait specifies period, duty factor, stride, clearance, six phase offsets, and per-leg stride and clearance scales. A damage-centered encoding maps the affected leg to a canonical index and reflects the phase relationships across sides. The shared candidate pool contains 24 gaits generated around the best intact gait at the current friction. The 24-candidate pool and ten-trial limit are development-stage budget choices, not optima established by a sensitivity sweep.

The development database contains 1,944 profiles from 81 conditions formed by crossing three damage types, legs 0--2, friction $\{0.90,1.00,1.10\}$, and onset $\{1.50,2.00,2.50\}$ s. Each condition uses twelve candidates and two seeds. A profile runs one candidate for twelve 0.75-s segments and a final 3-s interval. The eight targets are
\begin{equation}
\mathbf y=[v,q_E,q_M,q_L,q_F,d,p_D,e],
\label{eq:profile}
\end{equation}
where the targets are computed from twelve consecutive 0.75-s segments followed by a 3-s final execution of the same gait. Let $q_j$ be segment quality and $a_j,b_j$ its stable and torso--ground contact fractions. A stable time step has body tilt below $45^\circ$, trunk height above 0.09 m, and no torso--ground contact. The survival target is
\begin{equation}
v=\frac{1}{12}\sum_{j=1}^{12}\mathbf{1}\{a_j\geq0.8,\ b_j\leq0.2\}.
\end{equation}
Thus $v=1$ when all twelve segments meet these conditions and $v=0$ when none do; intermediate values are possible. It excludes the final interval and is distinct from the 0.95 stable-fraction gate used in Eq.~\ref{eq:quality}. The targets $q_E,q_M,q_L$ average $q_j$ over segments 1--4, 5--8, and 9--12, respectively. The target $q_F$ is quality in the subsequent 3-s interval. With $\bar q=12^{-1}\sum_jq_j$, define
\begin{equation}
\begin{split}
m&=\frac{\sum_{j=1}^{12}(j-6.5)(q_j-\bar q)}{\sum_{j=1}^{12}(j-6.5)^2},\\
d&=\operatorname{clip}_{[0,1]}\bigl(1+10\min(m,0)\bigr),\\
p_D&=\exp\bigl(-|y_F-y_{\mathrm{ref}}|/(1\,\mathrm{m})\bigr).
\end{split}
\end{equation}
Here $m$ is the least-squares quality slope per segment, $y_F$ is lateral position after the final interval, and $y_{\mathrm{ref}}$ is the fixed path reference. Despite its stored name \texttt{path\_debt\_score}, $p_D$ measures terminal path error, not accumulated error. The target $e$ is the final interval's normalized energy component from Eq.~\ref{eq:quality}, not energy averaged over the full profile. The online path penalty $\exp[-D/(0.08\,\mathrm{m})]$, defined below, instead uses positive increases in path error during a probe and extension and supplies a proxy correction to $p_D$.

For each target, five Gaussian-process regressors predict a mean and uncertainty from damage context and gait parameters \cite{rasmussen2006}. Members use seeded subsets of at most 192 examples and an RBF kernel. Predictive uncertainty combines posterior variance and ensemble disagreement. Grouped-condition cross-validation provides residuals for interval scaling to 95\% empirical coverage. This coverage is a calibration target on those residuals, not an independent coverage test.

\subsection{Conservative selection}
For candidate $i$ and target $j$, let $\mu_{ij}$ and $\sigma_{ij}$ denote the current mean and deviation. We rank candidates using lower estimates $L_{ij}=\mu_{ij}-\beta\sigma_{ij}$, with $\beta=0.50$ on the first trial and 0.35 thereafter. Durability utility is the geometric mean of clipped lower estimates for late quality, final quality, persistence, terminal path-error score, and energy, with each factor floored at $10^{-6}$.

On the first trial, candidates within 0.03 of the largest survival lower estimate are ranked by durability utility plus $0.01L_{iv}$. Later trials prioritize untried candidates with $L_{iv}\geq0.70$. If none meet this threshold, the policy chooses the untried candidate with the largest survival lower estimate. These estimates are selection heuristics, not calibrated survival-probability guarantees.

\subsection{Probe, update, and continue}
A probe is extended when transition, speed, and energy scores reach 0.20, 0.15, and 0.15, respectively. The observation updates nearby predictions through an RBF-weighted residual with weight 0.35 and reduces their uncertainty. Short-probe outcomes provide proxy updates to the long-horizon targets; they do not directly measure a new full profile.

We select the incumbent by combining observed performance with the predicted profile. A viable trial's observed score is the geometric mean of probe and extension quality, multiplied by $\exp(-D/0.08)$, where $D$ is the summed positive increase in global lateral error in metres. Its decision score multiplies this value by the square root of the geometric mean of predicted survival, late quality, final quality, and persistence. Nonviable trials receive zero observed score. The trial with the highest decision score supplies the bridge and final gait.

The robot executes a 0.75-s bridge after every trial. Adaptation stops when bridge quality falls below 0.35, falls below 60\% of the best bridge once that best exceeds 0.20, or follows two consecutive nonviable probes. It also stops at ten trials. The selected gait then runs for 3 s from the current state (Fig.~\ref{fig:method}). All coefficients and thresholds were fixed during development before holdout evaluation; no dedicated parameter sweep was conducted.

\section{Experimental Protocol}
The holdout crosses three damage types, previously unused legs 3--5, friction $\{0.865,1.135\}$, onset $\{1.73,2.37\}$ s, and three candidate seeds. This gives 36 physical conditions and 108 condition--seed tasks. The algorithm, surrogate, metrics, and evaluation procedure were fixed before opening this set. The reported holdout is now consumed and cannot serve as fresh validation after further tuning.

The three adaptive methods share the candidate pool and maximum trial budget. \emph{Random} starts with the best intact gait and samples the remaining candidates. \emph{PGRR} follows Section~IV. The remaining comparator is specified below so that its relation to IT\&E is explicit.

\subsection{IT\&E-inspired GP baseline}
The comparator uses a seven-dimensional gait descriptor comprising four contact-pattern features and three normalized gait parameters. The contact features are computed from 256 uniformly spaced phases of the commanded stance pattern and consist of support-count variation, left--right support imbalance, front--rear support imbalance, and tripod-stance imbalance. The remaining coordinates are normalized period, stride, and clearance,
\begin{equation}
\left(\frac{T-0.8}{1.6},\;\frac{S-0.03}{0.10},\;\frac{H-0.015}{0.055}\right).
\label{eq:ite_descriptor}
\end{equation}
The first four coordinates are, respectively, the standard deviation of total support count divided by 3, the mean absolute left--right count difference divided by 3, the mean absolute front--rear count difference divided by 2, and the mean absolute difference between the stance fractions of tripods $\{0,2,4\}$ and $\{1,3,5\}$. Unlike PGRR's input, this descriptor does not encode damage context or per-leg residual scales.

The GP uses the RBF kernel
\begin{equation}
k(\mathbf{x},\mathbf{x}')=0.08^2\exp\!\left(-\frac{\|\mathbf{x}-\mathbf{x}'\|^2}{2(0.5)^2}\right)
\label{eq:ite_kernel}
\end{equation}
and a friction-matched intact-quality prior mean $\boldsymbol{\mu}_0$. After observing gait indices $O$ with scores $\mathbf z$, the posterior mean is
\begin{equation}
\boldsymbol{\mu}^{+}=\boldsymbol{\mu}_0+K_{:,O}(K_{O,O}+10^{-3}I)^{-1}
(\mathbf z-\boldsymbol{\mu}_{0,O}),
\label{eq:ite_posterior}
\end{equation}
with uncertainty given by the standard GP posterior variance. The first candidate maximizes intact quality. Later candidates maximize $\mu_i^{+}+1.5\sigma_i$ over untried gaits, with seeded $10^{-10}$-scale Gaussian noise used only to break ties. The $10^{-3}$ term in Eq.~\ref{eq:ite_posterior} is observation regularization and is distinct from this tie-breaking jitter.

Each observation is the geometric mean of probe and extension quality multiplied by $\exp[-D/(0.08\,\mathrm{m})]$, or zero for a nonviable probe. The best observed score determines the bridge and final gait. The baseline shares PGRR's candidate pool, viability thresholds, segment durations, and maximum trial budget, but executes all ten trials without PGRR's early-stopping rule. Kernel and acquisition settings are fixed independently of PGRR. This implementation is a controlled IT\&E-inspired comparator rather than a reproduction of canonical IT\&E.

\begin{table}[H]
\caption{PGRR vs. IT\&E-inspired GP. Mean paired differences with 95\% bootstrap intervals.}
\label{tab:paired}
\centering
\renewcommand{\arraystretch}{1.15}

\resizebox{\columnwidth}{!}{%
\begin{tabular}{lccc}
\toprule

\textbf{Damage}
& \textbf{SRQ}
& \textbf{Final quality}
& \raisebox{-0.5\height}{\shortstack[c]{\textbf{Low-quality}\\\textbf{failure rate}}} \\

\midrule

All
& \shortstack{$+.044$\\$[+.014,+.075]$}
& \shortstack{$+.106$\\$[+.065,+.150]$}
& \shortstack{$-.278$\\$[-.370,-.185]$} \\

Amputation
& \shortstack{$+.027$\\$[-.031,+.083]$}
& \shortstack{$+.095$\\$[+.026,+.176]$}
& \shortstack{$-.222$\\$[-.389,-.056]$} \\

Fracture
& \shortstack{$+.023$\\$[-.020,+.068]$}
& \shortstack{$+.060$\\$[+.010,+.119]$}
& \shortstack{$-.194$\\$[-.361,-.056]$} \\

Joint lock
& \shortstack{$+.083$\\$[+.028,+.142]$}
& \shortstack{$+.164$\\$[+.078,+.256]$}
& \shortstack{$-.417$\\$[-.583,-.250]$} \\

\bottomrule
\end{tabular}%
}
\end{table}

\subsection{No adaptation and statistical analysis}
\emph{No adaptation} executes the default gait for ten 2.25-s segments and a 3-s final interval. It matches the maximum adaptive horizon, not the shorter duration of an early-stopped run. Its 36 deterministic condition-level results are reused across the three seed comparisons; they are not 108 independent runs.

We report means and paired method differences. Confidence intervals use 10,000 bootstrap resamples of the 108 matched condition--seed tasks; figure bands use 4,000 or 5,000 resamples. The intervals describe this task distribution, with possible dependence among seeds that share a physical condition. We therefore add a post-hoc sensitivity analysis using 10,000 paired condition-cluster bootstrap resamples. Each resample draws 36 physical conditions with replacement and retains all three matched seeds within each sampled condition. Damage-specific analyses resample the corresponding 12 conditions. We report percentile intervals without rerunning or tuning the method. This analysis addresses within-condition dependence but does not establish generalization to other robots or independent hardware replications.

\section{Results}
\subsection{PGRR improves final quality and reduces low-quality failures}
\begin{figure*}[!tb]
\centering
\includegraphics[width=\textwidth]{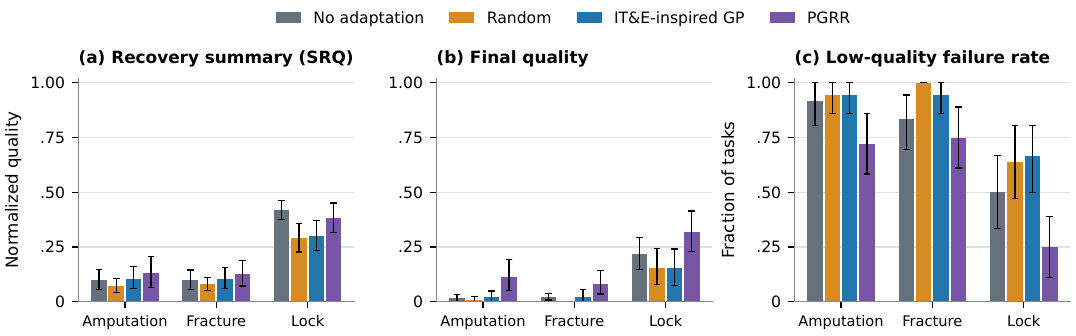}
\caption{Holdout outcomes for the four methods. Bars show means and 95\% task-bootstrap intervals. Higher is better for (a) SRQ and (b) final quality; lower is better for (c) low-quality failure rate ($q_F\leq0.05$). Each damage class has 36 condition--seed comparisons. No-adaptation results are reused across seeds. Amputation and fracture remain difficult in absolute terms.}
\label{fig:results}
\end{figure*}

Across the 108 tasks, PGRR improves SRQ by 0.044 and final quality by 0.106 relative to the IT\&E-inspired GP; the low-quality failure rate decreases by 0.278. All three paired intervals exclude zero (Table~\ref{tab:paired}). Final quality improves within amputation, fracture, and joint lock, with joint lock showing the largest paired gain. Figure~\ref{fig:results} shows that PGRR also has higher mean SRQ than the GP comparator in every damage class, although the amputation- and fracture-specific SRQ intervals include zero (Table~\ref{tab:paired}). Absolute scores remain low after amputation and fracture.

\subsection{Main gains persist under condition-cluster resampling}
Condition-cluster intervals exclude zero for all three aggregate differences against the IT\&E-inspired GP (Table~\ref{tab:cluster}). Final-quality gains also remain positive in each damage class. However, the damage-specific failure-rate intervals reach zero for amputation ($[-0.472,0.000]$) and fracture ($[-0.417,0.000]$). Evidence for reduced failures is therefore stronger in the pooled comparison and in joint lock than in those two classes individually. Against no adaptation, the clustered final-quality interval remains positive, whereas the SRQ interval includes zero. The main interpretation is unchanged when the three seeds sharing a physical condition are resampled together.

\begin{table}[!htb]
\caption{Condition-cluster sensitivity analysis with 95\% bootstrap intervals.}
\label{tab:cluster}
\centering
\renewcommand{\arraystretch}{1.15}

\resizebox{\columnwidth}{!}{%
\begin{tabular}{lcc}
\toprule

\textbf{Outcome}
& \textbf{IT\&E-inspired GP}
& \textbf{No adaptation} \\

\midrule

SRQ
& $+.044\,[+.008,+.082]$
& $+.009\,[-.018,+.037]$ \\

Final quality
& $+.106\,[+.062,+.153]$
& $+.086\,[+.046,+.131]$ \\

Failure rate
& $-.278\,[-.398,-.167]$
& $-.176\,[-.306,-.065]$ \\

\bottomrule
\end{tabular}%
}
\end{table}

\subsection{Final-gait gains are not captured by the adaptation summary alone}
Against no adaptation, PGRR improves final quality by 0.086 [0.049, 0.125] and reduces low-quality failures by 0.176 [0.093, 0.259]. The aggregate SRQ difference is only +0.009 [-0.017, 0.035]. Damage-specific means explain this result. PGRR raises SRQ by 0.035 after amputation and 0.027 after fracture, equivalent to 36\% and 28\% of the corresponding no-adaptation means, although both paired intervals include zero. For joint lock, SRQ decreases by 0.035 [-0.090, 0.020], even as final quality improves by 0.098 [0.021, 0.177].

The no-adaptation controller already retains substantially more recovery-summary quality under joint lock (0.418) than under amputation (0.097) or fracture (0.099). PGRR also executes a mean of 4.0 trials for joint lock, compared with 1.86 for amputation and 1.67 for fracture. Because SRQ includes the quality of probes, extensions, and bridges, this pattern is consistent with adaptation costs offsetting the benefit of a better final gait when the unchanged controller remains comparatively effective. It does not show that elapsed search time alone causes the decrease, because SRQ is a slot average rather than a wall-clock integral. The result instead demonstrates that endpoint improvement and recovery-summary improvement need not coincide.

\begin{figure*}[!tb]
\centering
\includegraphics[width=\textwidth]
{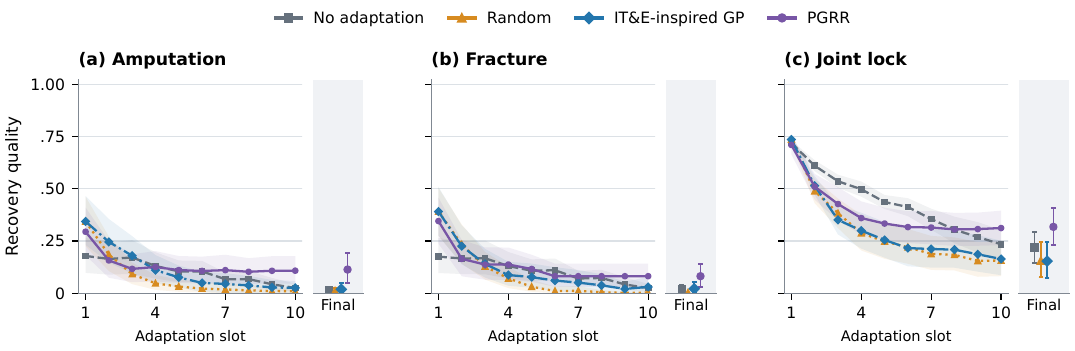}
\vspace{-20pt}
\caption{Post-hoc recovery-slot profiles (mean and 95\% task-bootstrap interval). Slots average executed probe, extension, and bridge segments; slots after early stopping are filled with final quality. The shaded final region shows the separate 3-s endpoint, with small horizontal offsets for readability. The x-axis is not elapsed time. Padded tails are not additional measurements of sustained walking.}
\label{fig:trajectory}
\end{figure*}

Figure~\ref{fig:trajectory} shows the post-hoc slot profile. PGRR has the highest mean final quality among adaptive methods in all three damage classes. However, the plotted slots combine executed trials with final-quality padding after early stopping. PGRR stops after its first trial in 25 of 36 amputation tasks, 25 of 36 fracture tasks, and 2 of 36 joint-lock tasks. Flat tails therefore do not demonstrate continuous physical persistence. The plot describes how the reported score is formed, rather than a measured wall-clock recovery trajectory.

\subsection{Removing the uncertainty penalty reduces performance}
We compare three variants on 27 matched development tasks. The \emph{myopic profile} variant ranks early quality while retaining the survival constraint. The \emph{mean acquisition} variant removes the uncertainty penalty, and the \emph{default bridge} variant uses the default gait between probes. Table~\ref{tab:ablation} reports variant-minus-full differences. Removing the uncertainty penalty lowers SRQ and final quality, with both intervals excluding zero. The profile-ranking and bridge contrasts are negative on average but inconclusive. These data support conservative acquisition more clearly than an independent benefit from long-horizon ranking. The conditions overlap surrogate training, so this is a development diagnostic rather than an independent mechanism test.

\begin{table}[!t]
\caption{Development ablations. Mean variant-minus-PGRR differences with 95\% bootstrap intervals.}
\label{tab:ablation}
\centering
\renewcommand{\arraystretch}{1.15}

\resizebox{\columnwidth}{!}{%
\begin{tabular}{lccc}
\toprule

\textbf{Variant}
& \textbf{SRQ}
& \textbf{Final quality}
& \raisebox{-0.5\height}{\shortstack[c]{\textbf{Low-quality}\\\textbf{failure rate}}} \\

\midrule

Myopic profile
& \shortstack{$-.046$\\$[-.096,+.001]$}
& \shortstack{$-.041$\\$[-.127,+.055]$}
& \shortstack{$+.037$\\$[-.074,+.185]$} \\

\shortstack[c]{Mean acquisition\\($\beta=0$)}
& \shortstack{$-.045$\\$[-.083,-.011]$}
& \shortstack{$-.068$\\$[-.136,-.011]$}
& \shortstack{$+.037$\\$[+.000,+.111]$} \\

Default bridge
& \shortstack{$-.035$\\$[-.092,+.020]$}
& \shortstack{$-.039$\\$[-.117,+.041]$}
& \shortstack{$+.000$\\$[-.148,+.148]$} \\

\bottomrule
\end{tabular}%
}
\end{table}

Prediction accuracy also varies across profile targets. Grouped-condition prediction correlations are 0.879 for survival, 0.830/0.795/0.713 for early/middle/late quality, 0.651 for final quality, 0.382 for persistence, 0.610 for terminal path-error score, and 0.691 for energy. Mean absolute errors range from 0.124 to 0.192. Weak persistence prediction and the inconclusive myopic ablation limit claims that the long-horizon profile itself explains the gains.

\section{Discussion}

The clearest finding is the distinction between final gait selection and the recovery summary. PGRR improves final quality against both the IT\&E-inspired GP and no adaptation, whereas its aggregate SRQ advantage is conclusive only against the GP. Under joint lock, the default gait remains comparatively effective, so the probe--extension--bridge sequence can lower the recovery summary even when PGRR identifies a better final gait. Under amputation and fracture, frequent first-trial stopping means that good initial selection may account for a substantial part of the benefit. SRQ also includes final quality and fills unused slots after early stopping; it therefore does not isolate motion during exploration.

Development ablations identify conservative uncertainty handling as the best-supported component: removing its penalty reduces both SRQ and final quality. The profile-ranking and bridge contrasts remain inconclusive, and persistence is the least accurately predicted target. These results support the combined PGRR policy while leaving the independent contribution of repeated-execution prediction less certain.

\section{Conclusion}
We introduced PGRR, a reset-free damage-recovery framework that combines repeated-execution profiles, conservative uncertainty-aware selection, and online correction. Short probes, conditional extensions, and incumbent bridges preserve a continuous physical trajectory, while a separate 3-s interval measures the selected gait from the state reached at the end of adaptation.

Across 108 held-out MuJoCo tasks, PGRR improves final quality by 0.106 and SRQ by 0.044 relative to the tested IT\&E-inspired GP, while reducing low-quality failures by 27.8 percentage points. Final-quality gains occur after amputation, fracture, and joint lock, and the aggregate results remain positive when bootstrap resampling groups seeds by physical condition. Against no adaptation, PGRR improves final quality and reduces failures, but the aggregate SRQ difference is inconclusive. Joint lock makes the distinction concrete: the unchanged gait remains comparatively effective, so adaptation trials can reduce SRQ even when the selected final gait improves.

These findings show that final-gait quality and the recovery summary capture complementary aspects of reset-free recovery and should be reported together. Development ablations most clearly support conservative uncertainty handling. PGRR provides a practical framework for improving and evaluating continuous-state post-damage recovery.


\end{document}